\documentclass[sigconf]{acmart}
\usepackage{url}
\usepackage{float}
\usepackage{placeins}
\usepackage{ulem}
\usepackage{amsmath,amstext} 
\usepackage{stfloats}
\usepackage[toc,page]{appendix}
\graphicspath{ {figs/} }
\setcopyright{acmlicensed}
\copyrightyear{2019}
\acmYear{2019}
\acmConference[CASCON '19]{CASCON '19}{November 4-6, 2019}{Toronto, Canada}
\acmPrice{}
\acmDOI{}
\acmISBN{}

\title{Predicting student performance using data from an Auto-grading system}

\author{Huanyi Chen} \affiliation{\institution{University of
    Waterloo}\city{Waterloo, ON}\country{Canada}}
\email{huanyi.chen@uwaterloo.ca}

\author{Paul A.S. Ward} \affiliation{\institution{University of
    Waterloo}\city{Waterloo, ON}\country{Canada}} \email{pasward@uwaterloo.ca}

\begin{document}

\newcommand{\firstplace}{\textbf}
\newcommand{\secondplace}{\emph}

%
\begin{abstract}
  As online auto-grading systems appear, information obtained from those systems
  can potentially enable researchers to create predictive models to predict
  student behaviour and performances. In the University of Waterloo, the ECE 150
  (Fundamentals of Programming) Instructional Team wants to get an insight into
  how to allocate the limited teaching resources better to achieve improved
  educational outcomes. Currently, the Instructional Team allocates tutoring
  time in a reactive basis. They help students ``as-requested''. This approach
  serves those students with the wherewithal to request help; however, many of
  the students who are struggling do not reach out for assistance.
  Therefore, we, as the Research Team, want to explore if we can determine
  students which need help by looking into the data from our auto-grading system,
  Marmoset.

  In this paper, we conducted experiments building decision-tree and
  linear-regression models with various features extracted from the Marmoset
  auto-grading system, including passing rate, testcase outcomes, number of
  submissions and submission time intervals (the time interval between the
  student's first reasonable submission and the deadline). For each feature, we
  interpreted the result at the confusion matrix level. Specifically for
  poor-performance students, we show that the linear-regression model using
  submission time intervals performs the best among all models in terms of
  Precision and F-Measure. We also show that for students who are misclassified
  into poor-performance students, they have the lowest actual grades in the
  linear-regression model among all models. In addition, we show that for the
  midterm, the submission time interval of the last assignment before the
  midterm predicts the midterm performance the most. However, for the final exam,
  the midterm performance contributes the most on the final exam performance.

\end{abstract}

%
%


%

%
\maketitle

\section{Introduction}
\label{sec:introduction}

In programming courses, if students can get feedback in a short time 
after they submit their code, they might be able to improve quickly. However, it
is not realistic for instructors or teaching assistants (TAs) to mark students'
code repeatedly and generate feedback if students can submit their code
multiple times. The reason is that students may potentially have a large amount of
submissions, and it may take much time for Instructional Team to mark before they
can give feedback. Thus, an auto-grading system is needed. A sound auto-grading
system can not only give out the testcase outcomes but also generate useful
feedback for the students. Since the system can keep running in the background,
the feedback can be quickly generated whenever students submit their code
(within a few minutes, typically).

The auto-grading system, Marmoset, allows students to submit code multiple times
for a coding assignment. Marmoset then will test the code and
record the testcase outcomes into a MySQL database~\cite{spacco-msr2005}.
Students can keep modifying their code until they get an acceptable score, or
until the deadline. Therefore, assignment marks or testcase outcomes may not
necessarily be a valuable metric to judge if a student has learned the material
well or not. However, their behaviour, such as how frequently they submit, how
early they submit for the first time, and so on, might give valuable
information.

In our study, we, as the Research Team, are focusing on investigating the
following research themes.

\begin{enumerate}
\item If we put students into categories according to their final exam and
  midterm performances, can we create a model over the Marmoset data to
  understand the students' behaviour and predict those categories?
\item Can we predict the students' numerical midterm grades and the final exam
  grades using the Marmoset data?
\item Can we find any interesting relations between the Marmoset data, grades
  and student categories?
\end{enumerate}

\section{Related Work}
\label{sec:related_work}

Many studies were focusing on predicting the student's overall
performance. \citet{nghe2007comparative} tried to predict the student's actual
GPA at the end of the year using the student's information from the previous year.
\citet{dekker2009predicting} described case studies of electrical engineering
students to predict if they will drop out program after the first semester of
their studies, or even before they enter the study program.
\citet{khobragade2015students} conducted several experiments using students data
to predict if a student will fail or not at the end of the semester. These
studies usually use data from a wide range of years and a wide range of courses.

For programming courses, \citet{koprinska2015students} tried to predict whether
students will pass or fail their final exam with three different sources:
progression in code writing and diagnostic from an auto-grading system (PASTA),
interaction and engagement from online discussion boards and wikis (Piazza), and
assessment marks. \citet{gramoli2016mining} studied the contribution of auto-grading
and instant feedback using the PASTA system in various computer science courses.
However, the only student performance prediction mentioned in their paper is
that the students who start submitting their work early or finish submitting
early (meaning that they finished the work) tend to get higher marks, but no
complete evaluation provided. However, their data is not within one semester, and
auto-grading data is not the primary data source.

The few studies in which the researchers only consider the data from an
auto-grading system as the data source, however, were not trying to use the data
for student performance prediction. For example, \citet{mcbroom2016mining} mined
data from the PASTA system and cluster the students according to their
submissions. However, rather than making a suggestive prediction on an
individual student for the Instructional Teams, they focused on analyzing
all students' behaviour and the evolution of the behaviour over the
semester.

Because of the easy-collecting nature of the data from auto-grading systems,
predictive models using these data can be easily integrated into other
educational systems. We believe such predictive models can further help the
Instructional Teams to allocate their limited teaching resources on students.

\section{Student Performance}

In the University of Waterloo, student performance for one course is usually
calculated based on a mixture of assignments' marks, projects' marks, midterm
grades, and final exam grades. In terms of the course Fundamentals of
Programming (ECE 150) in Electrical and Computer Engineering department, the
midterm grades and final exam grades contribute from $60\%$ to $90\%$ to the
final course grades. In other words, student performance highly depends on the
midterm grades and final exam grades. In the year 2016, the maximum score for
the midterm for ECE 150 was 110, and for the final exam, it was 120.

\section{Marmoset}

Marmoset~\cite{marmoset:Spacco:2006:MPA:1176617.1176665} is an auto-grading
system for marking programming project submissions. It was built at and used
for, computer science courses in the University of Maryland. In the year 2015,
it was integrated as the auto-grading tool for ECE 150 (Fundamentals of
Programming) and ECE 356 (Database systems) in the University of Waterloo.

In ECE 150, an assignment usually comes with multiple tasks. Each task
corresponds to a project in Marmoset and contains several testcases. There are
three types of testcases: public testcases, release testcases, and secret
testcases.

\begin{figure}[htb]
  \centering \includegraphics[width=\columnwidth]{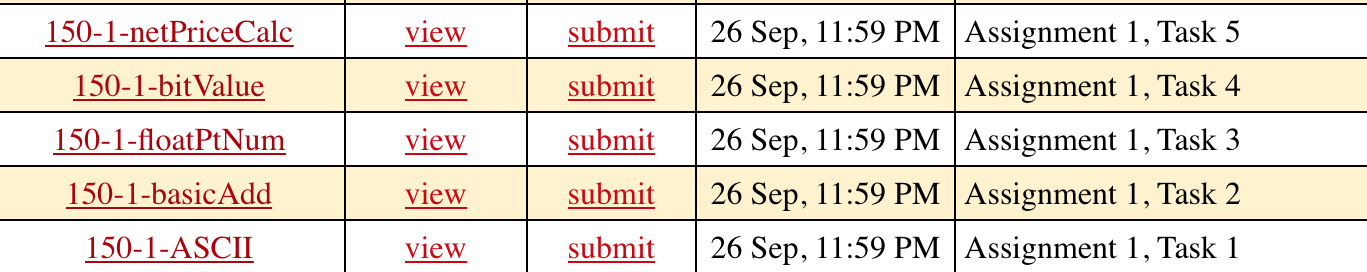}
  \caption{The graded tasks in Marmoset}
  \label{fig:marmoset_graded_tasks_list}
\end{figure}

\begin{figure}[htb]
  \centering \includegraphics[width=\columnwidth]{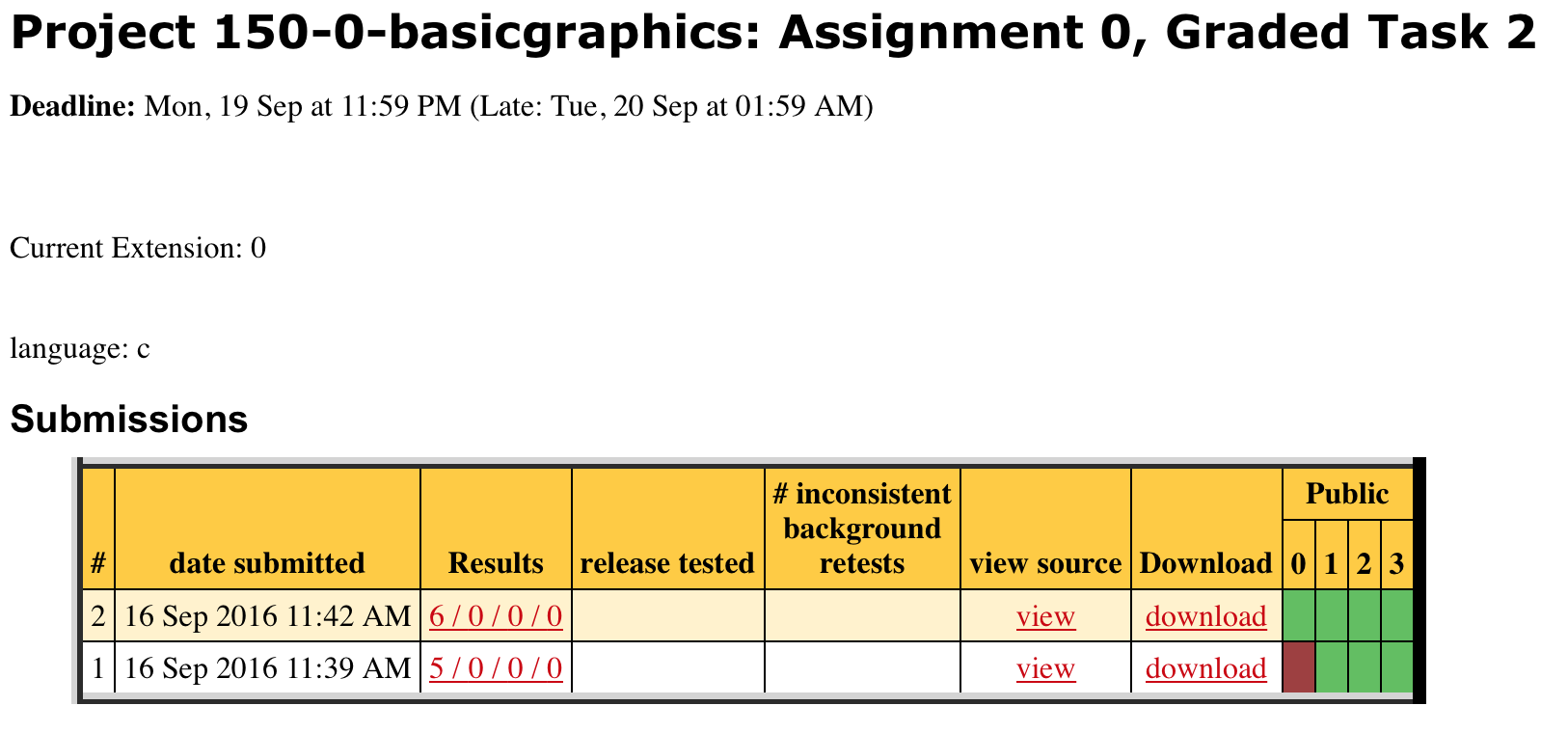}
  \caption{Student view of a task in Marmoset}
  \label{fig:marmoset_task_test_results}
\end{figure}

\begin{figure}[htb]
  \centering \includegraphics[width=\columnwidth]{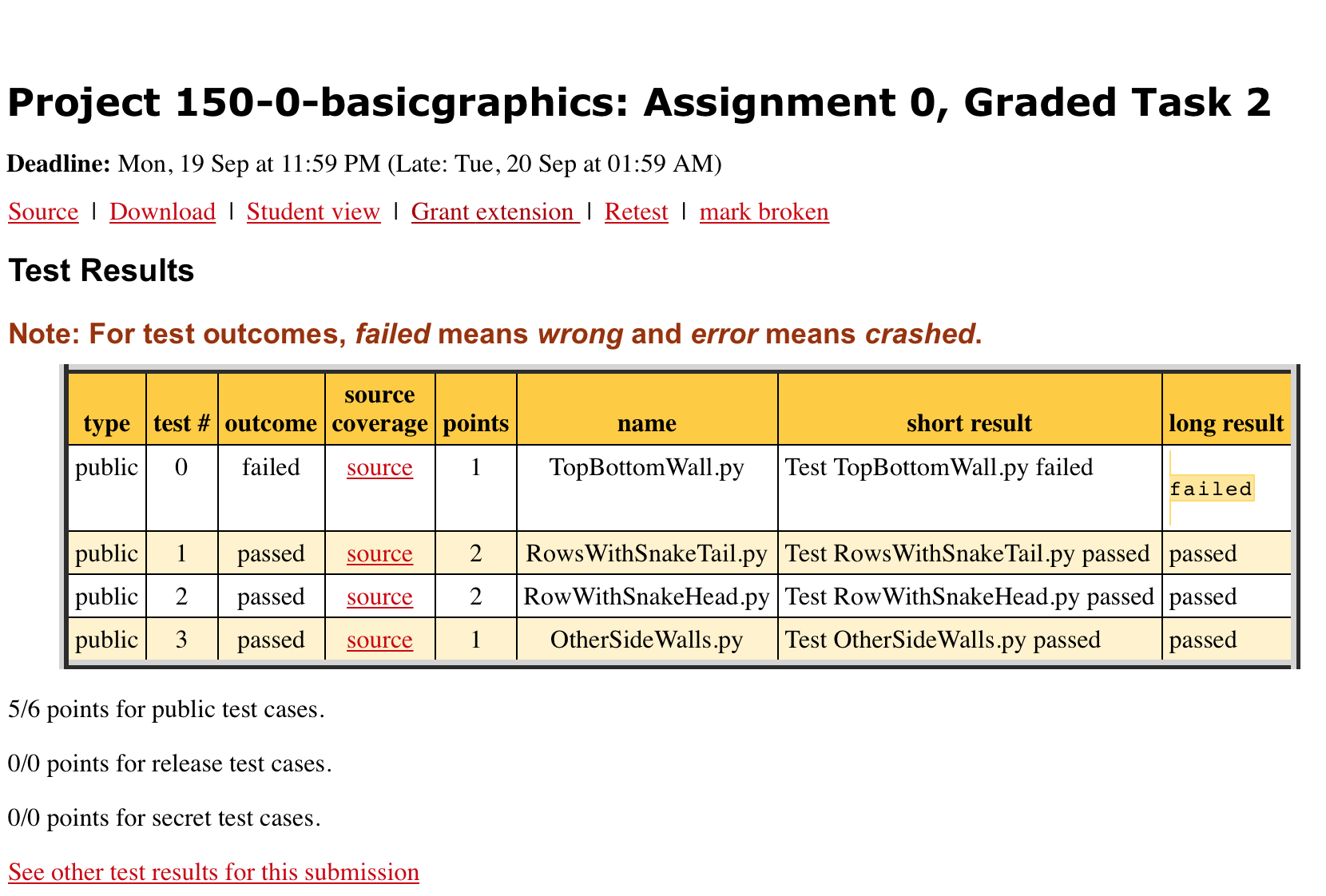}
  \caption{Test results of a task in Marmoset}
  \label{fig:marmoset_task_test_result_details}
\end{figure}

The naming convention for the year 2016 of ECE 150 is ``course
number''-``assignment number''-``task name'' for graded tasks.
Figure~\ref{fig:marmoset_graded_tasks_list} shows the graded tasks for
assignment 1 of ECE 150. Notice that they have the same deadline.
Figure~\ref{fig:marmoset_task_test_results} shows us the student view of the
test results in Marmoset. In the year 2016, the Instructional Team only used
public testcases (shown in the right part of the figure) but no release
testcases nor secret testcases. Green colour means a submission passed a given
testcase while red colour means it failed that testcase. There is also grey colour
meaning the submission failed to compile. Points for testcases can be different.
Each testcase may not always worth one point. Therefore, the top submission in
figure~\ref{fig:marmoset_task_test_results} gets a result as ``6/0/0/0''
because testcases were worth different points.
Figure~\ref{fig:marmoset_task_test_result_details} shows us the student view of
the detailed test results and feedback for the second submission in
figure~\ref{fig:marmoset_task_test_results}. Students can pick any task to
start, and they can switch between tasks even if they have not passed all testcases.

\section{Modelling Techniques}
\label{sec:modelling_techniques}

This section describes the modelling techniques used in predicting student
performance. We selected decision tree and linear regression in our experiments.

\subsection{Decision Tree}

Decision tree is a method for approximating discrete-valued target functions, in
which the learned function is represented by a decision tree. Decision trees
classify instances by sorting them down the tree from the root to a leaf node,
which provides the classification of the instance. Each node in the tree
specifies a test of some attribute of the instance, and each branch descending
from that node corresponds to one of the possible values for this attribute. An
instance is classified by starting at the root node of the tree, testing the
attribute specified by this node, then moving down the tree branch corresponding
to the value of the attribute. This process is then repeated for the subtree
rooted at the new node.

\paragraph{C4.5 Decision Tree~\cite{Quinlan1993}} C4.5 builds decision trees
from a set of training data by choosing the split that gives the maximum
information gain ratio using the concept of information entropy. The attribute
with the highest normalized information gain is chosen to make the decision.
C4.5 is also able to handle both continuous and discrete attributes. In
Weka~\cite{WEKA:Witten:2016:DMF:3086818}, the class implements this is called
J48. Many researchers use this technique for exploring their
data~\cite{nghe2007comparative, dekker2009predicting, bydvzovska2016comparative,
  costa2017evaluating}, and this is the decision tree algorithm used in our
experiments.

\subsection{Linear Regression}

Linear regression assumes that the relationship between a dependent variable and
an independent variable can be described in a linear manner. The target is to
find the best estimation of the value for the coefficient of the equation by
using the values of the independent variable. A standard approach in finding the
coefficient is applying the least-squares method.

The error between observation and the true value can be caused by measurement
error, system error, etc. It can be thought of as the composite of a number of
minor influences or errors. As the number of these minor influences gets large,
the distribution of the error tends to approach the normal distribution (Central
Limit Theorem~\cite{le1986central}). Therefore, the error follows:
$$\epsilon \sim N(0,\sigma^2)$$

It should be equal variance and normally distributed.

\paragraph{Multiple Linear Regression}

The difference between linear regression and multiple linear regression is that
in linear regression, we have only one independent variable; however, in
multiple linear regression, we can have multiple independent variables. We will
refer to linear regression as multiple linear regression in later sections.

\subsubsection{Correlation Coefficient}

The correlation coefficient measures the degree to which two variables are
related in a linear relationship. A positive relationship between two variables
means if there is a positive increase in one variable, there is also a positive
increase in the second variable. A value of precisely $1.0$ means a perfect
positive relationship. In comparison, a value of precisely $-1.0$ means a perfect
negative relationship, which means the two variables move in opposite
directions. However, if the correlation is $0$, it means no relationship exists
between the two variables. The value of the correlation coefficient indicates
the strength of the relationship. For example, a value of $0.3$ indicates a very
weak positive relationship.

\subsubsection{Verify linear regression assumptions}

In order to verify the linear regression assumptions that the error is equal variance and
normally distributed, graphic techniques are often applied. A residual vs fitted
plot is used for visually observing non-linearity, unequal error variances, and
outliers~\cite{residual_vs_fitted}. A normal Q-Q plot is the plot to estimate if
a dataset is from theoretical normal distribution~\cite{normal_qq}. A residuals
vs leverage plot illustrates influential points. Not all outliers are
influential in the linear-regression analysis. Even though data have extreme values,
they might not be influential in determining a regression line. However, some
points could be very influential even if they appear to be within a reasonable
range of values. They could be extreme points against a regression line and
can alter the results if we exclude them from the analysis.

\section{Features}

This section describes the features.

\subsection{Passing Rate for each task} This feature reflects the passing rate
of the best submission of a student for a given task. In the course ECE 150
during the year 2016, the total number of graded tasks for all students is 28.
Among these tasks, the first 16 tasks are due before the midterm, and the other
12 tasks are released after midterm and due before the final exam. Note that
multiple tasks can belong to one assignment, which means they have the same
deadline.

\begin{figure}[htb]
  \centering
  \includegraphics[width=\columnwidth]{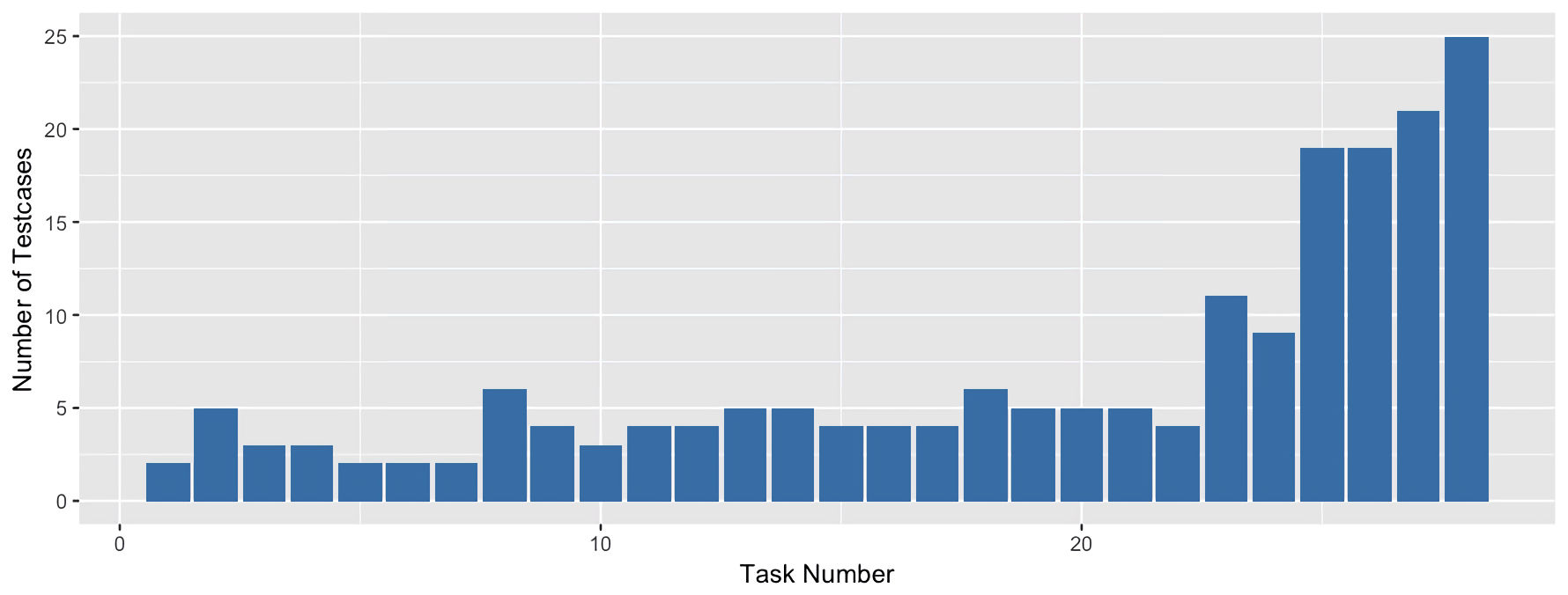}
  \caption{Number of testcases of tasks}
  \label{fig:num_of_testcases}
\end{figure}

The number of testcases differs from one task to another. We show the distribution of
testcases in Figure \ref{fig:num_of_testcases}. The larger the task
number is, the closer towards the final exam.

If a task has $n$ testcases, and the best submission of a student passes $m$
testcases, then the passing rate is calculated by $m/n$. We expect this feature
to be of limited value in prediction because students can keep submitting until
they pass all the testcases.

\subsection{Testcase Outcomes} This feature is similar to the passing rate
feature. It reflects the testcase outcomes of the best submission of a student
for a given task. However, different from the passing rate, this feature
emphasizes on individual testcase. It provides us with an insight into the contribution of each
testcase and might be able to tell us which testcases are important.

\subsection{Submission Time Interval} The time interval between the time of
submissions and the task deadline might be considered as a useful feature. In
Gramoli et al.~\cite{gramoli2016mining} study, they observe that students who
start submitting early or stop (re)submitting early to the auto-grading system
tend to get higher marks in their assignments.

In our experiments, we consider using the time interval between the time of the
submission passing 75\% testcases and the task deadline as the feature.

\begin{figure}[htb]
  \centering
  \includegraphics[width=\columnwidth]{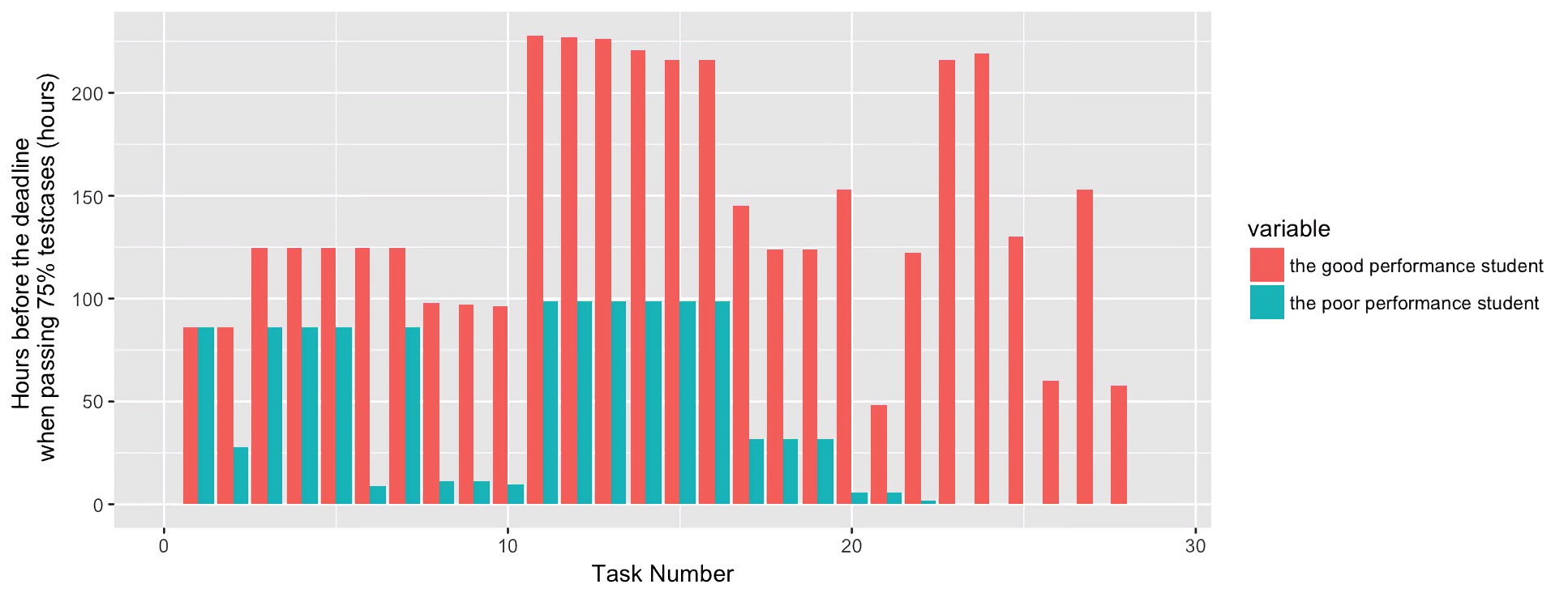}
  \caption{Time interval difference between a good-performance student and a
    poor-performance student}
  \label{fig:subtime_examples_bar}
\end{figure}

Figure \ref{fig:subtime_examples_bar} shows us an example of the feature, the
time interval difference between a randomly selected poor-performance student
(32/110 for the midterm and 45/120 for the final exam) and a randomly selected
good-performance student (109/110 for the midterm and 103/120 for the final exam).
Here the Y-axis shows us the number of hours before the deadline when the
student makes a submission which passes at least $75\%$ of the testcases for a
given task. Note that the average hours of poor-performance students for all
tasks reside mostly between 10 hours and 60 hours while the average hours of
good-performance students reside mostly between 100 hours and 140 hours. The
difference is quite significant.

\subsection{Number of Submissions} The number of submission feature can be
directly generated from the number of submission information stored in the MySQL
database of Marmoset. Typically, the last submission is the best.
The reason is that it is common that a student stops working on the task if he
has already passed all the testcases; if he has not achieved that, he might keep
working on it. It means the last submission will be the best. For a
smart student, who may improve his/her code significantly from the feedbacks for
his/her previous submission, the number of submissions should be relatively
small. However, for a less smart student, he may need to make more
submissions to get his code correct. Therefore, the number of submissions might
be able to help us predict student performance.

\begin{figure}[htb]
  \centering \includegraphics[width=\columnwidth]{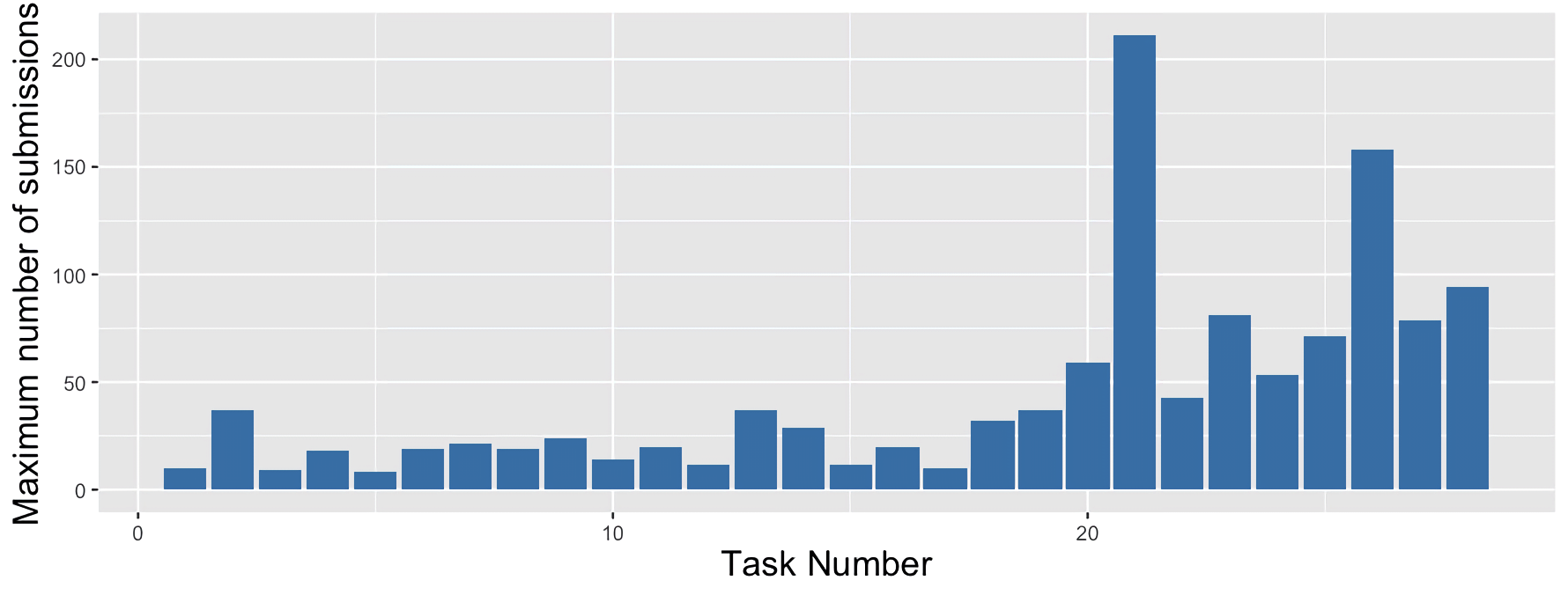}
  \caption{Maximum number of submissions}
  \label{fig:nsub_max}
\end{figure}

Figure \ref{fig:nsub_max} gives us the maximum number of submissions. We notice
that the maximum number of submissions can be huge, such as 211 times (from the
$21^{st}$ task). Also, for the tasks after the midterm (the last 12 tasks), it
becomes more substantial than those before the midterm (the first 16 tasks).

\section{Experiments and Evaluation}
\label{sec:experiments_and_evaluation}

This section describes the experiments and evaluation of our study.
Weka~\cite{WEKA:Witten:2016:DMF:3086818} is a tool containing many machine
learning algorithms for data mining tasks, and it is the tool we used in the
experiments.

\subsection{Methodology}

We conduct two types of experiments: classification to predict student
categories; regression to predict the grades of the midterm and the final exam.

The dataset in our experiments contains 428 instances which representing the
students who have both the midterm grade and the final exam grade in ECE 150 for
the Fall term in the year 2016. Students who missed either the midterm exam or
the final exam are excluded.

We split the dataset into a training set and a testing set. We use the training
set to build the model; then we apply the model to the testing set for
evaluation. We use Weka to conduct the experiments, and the default parameter
settings of the algorithms in Weka are used. For the classification method, we
select the C4.5 decision-tree algorithm; for the regression method, we select
the multiple linear regression algorithm.

We refer to good-performance students as \textbf{GP}, satisfactory-performance
students as \textbf{SP}, and poor-performance students as \textbf{PP} in the
following tables.

\subsection{Classification for Predicting Student Categories}

This section describes the classification experiments for predicting the student
categories aiming to explore the answer of the first research question: If we
put students into categories according to their midterm and final exam
performances, can we create a model over the auto-grading data to understand the
students' behaviour and predict those categories?

In this section, we describe our experiments for predicting student categories
using the C4.5 decision-tree classification algorithm and their results. We
split the students into three categories according to their exam grades. We apply
the same rule for both the midterm and the final exam: Students who get higher
than 80 are categorized as good-performance students; students who get lower
than 50 are categorized as poor-performance students; the remaining students are
categorized as satisfactory-performance students. Note that students who are
categorized in one category for the midterm can be in a different category for the
final exam. Here, the number $80$ is selected according to common sense, the
number $50$ is the term average for not withdrawing the engineering program in
University of Waterloo. For predicting the midterm performance, we only use the
tasks assigned before the midterm (16 tasks). For predicting the final exam
performance, we use all the tasks (28 tasks).

In the experiments, the training set to testing set ratio is 8:2 for both the
midterm and the final exam. We apply the Synthetic Minority Over-sampling
Technique (SMOTE)~\cite{Chawla:2002:SSM:1622407.1622416} to oversample the poor
performance students in both training sets for building better predictive
models.

\subsubsection{Passing Rate as the Feature}

For each submission, Marmoset tests the code against pre-set testcases, and
stores the testcase outcomes. For each testcase, the test outcome will be either
`passed' or `failed'. The passing rate for a task is calculated by the number of
testcases passed divided by the total number of testcases for that task. The
experiment result for classifying students according to their midterm grades is
shown in Table \ref{table:passing_rate_for_stu_categories_cm_midterm}. The
experiment result for classifying students according to their final exam grades
is shown in Table \ref{table:passing_rate_for_stu_categories_cm_final}.

\begin{table}[htb]
  \centering
  \begin{tabular}{cccc}
    \hline
    \hline
    PP & SP&GP&$\leftarrow$ classified as\\
    0               & 0                      &10                          & PP\\
    2               & 3                      &27                          & SP\\
    0               & 2                      &42                          & GP\\
    \hline
  \end{tabular}
  \caption{Confusion Matrix for Passing Rate (midterm)}
  \label{table:passing_rate_for_stu_categories_cm_midterm}
\end{table}
\begin{table}[htb]
  \centering
  \begin{tabular}{cccc}
    \hline
    \hline
    PP&SP&GP&$\leftarrow$ classified as\\
    2               & 2                      & 9                          & PP\\
    3               & 6                      &18                          & SP\\
    0               & 9                      &37                          & GP\\
    \hline
  \end{tabular}
  \caption{Confusion Matrix for Passing Rate (final)}
  \label{table:passing_rate_for_stu_categories_cm_final}
\end{table}

From table \ref{table:passing_rate_for_stu_categories_cm_midterm} and table
\ref{table:passing_rate_for_stu_categories_cm_final}, we can tell in both
models, most of the students are classified as good-performance students. In terms
of the poor-performance students (whom we are more care about), the midterm
model makes no correct prediction for them. Even worse, all poor-performance
students are classified as good-performance students. However, it turns better
when it comes to the final exam, which classifies some number of poor-performance
students correctly. It might indicate the passing rate for tasks after midterm
are able to reflect some variance among students since the tasks after midterm
might be more difficult. We will discuss more the comparison in terms of
poor-performance students in later sections.

\subsubsection{Testcase Outcomes as the Feature}

Different from the passing-rate feature, an individual testcase outcome from
the best submissions might be useful. A poor-performance student might satisfy
with the overall passing rate for the task, but not passing all the testcases.
Therefore, the individual testcases might help us to build a useful model.
Specifically, certain testcase may act as differentiation between students.
The experiment result for classifying students according to their midterm grades
is shown in Table \ref{table:testcases_for_stu_categories_cm_midterm}. The
experiment result for classifying students according to their final exam grades
is shown in Table \ref{table:testcases_for_stu_categories_cm_final}.

\begin{table}[htb]
  \centering
  \begin{tabular}{cccc}
    \hline
    \hline
    PP&SP&GP&$\leftarrow$ classified as\\
    0               & 1                      &9              & PP \\
    0               & 5                      &27             & SP \\
    0               & 6                      &40             & GP\\
    \hline
  \end{tabular}
  \caption{Confusion Matrix for Testcases Outcomes (midterm)}
  \label{table:testcases_for_stu_categories_cm_midterm}
\end{table}
\begin{table}[htb]
  \centering
  \begin{tabular}{cccc}
    \hline
    \hline
    PP&SP&GP&$\leftarrow$ classified as\\
    1               & 4                      &8              & PP\\
    2               &11                      &14             & SP\\
    1               & 8                      &37             & GP\\
    \hline
  \end{tabular}
  \caption{Confusion Matrix for Testcases Outcomes (final)}
  \label{table:testcases_for_stu_categories_cm_final}
\end{table}

From table \ref{table:testcases_for_stu_categories_cm_midterm} and table
\ref{table:testcases_for_stu_categories_cm_final}, unexpectedly, the results are
very similar to the results using passing-rate as the feature. The students are
biased to be predicted as good-performance students. It indicates that there are
no outstanding testcases which can be used to differentiate students. Since it
is an introduction level programming course, the testcases are relatively straightforward, so
it makes sense that there are no outstanding testcases.

\subsubsection{Number of Submissions as the Feature}

The experiment result for classifying students according to midterm grades is
shown in Table \ref{table:abs_num_submission_for_stu_categories_cm_midterm}. The
experiment result for classifying students according to their final grades is
shown in Table \ref{table:abs_num_submission_for_stu_categories_cm_final}.

\begin{table}[htb]
  \centering
  \begin{tabular}{cccc}
    \hline
    \hline
    PP&SP&GP&$\leftarrow$ classified as\\
    3               & 3                      & 4              & PP\\
    6               &13                      &13              & SP\\
    6               &17                      &21              & GP\\
    \hline
  \end{tabular}
  \caption{Confusion Matrix for Number of Submissions (midterm)}
  \label{table:abs_num_submission_for_stu_categories_cm_midterm}
\end{table}

\begin{table}[htb]
  \centering
  \begin{tabular}{cccc}
    \hline
    \hline
    PP&SP&GP&$\leftarrow$ classified as\\
    7               & 5                      &1               & PP\\
    1               &10                      &16              & SP\\
    6               &14                      &26              & GP\\
    \hline
  \end{tabular}
  \caption{Confusion Matrix for Number of Submissions (final)}
  \label{table:abs_num_submission_for_stu_categories_cm_final}
\end{table}

From table \ref{table:abs_num_submission_for_stu_categories_cm_midterm} and
table \ref{table:abs_num_submission_for_stu_categories_cm_final}, it seems both
models manage to make some predictions of the poor-performance students.
However, for the midterm, most of the predicted poor-performance students are
satisfactory students and good-performance students (12 out of 15). For
the final exam, most of the predicted poor-performance students are
poor-performance students and good-performance students. Because there is more considerable
variance in this feature after the midterm, as shown in figure
\ref{fig:nsub_max}, the result is expected. It might indicate for the midterm,
some satisfactory students and some good-performance students behave similarly
as poor-performance students. However, for the final exam, only a small fraction
of satisfactory students will behave like poor-performance students, while for
good-performance students, there are still some good-performance students who
behave similarly as poor-performance students, which indicates that the number
of submissions feature itself cannot differentiate good-performance students
from poor-performance students. Combination of some other features might be
helpful.

\subsubsection{Time Interval between certain Submission and Deadline as the
  Feature}
\label{subsection:submission_time_interval}

The feature in our experiment is the time interval between the student's
submission time and the deadline of the project. We calculate how many hours
before the deadline, a student submits as the feature (a submission time after
the deadline or no submission, is deemed as 0 hour before the deadline). Then we
try to build a decision tree using the submission time intervals to predict the
students' categories.

Instead of using the submission time of the last submission, we used the
submission time of the submission when it gets $75\%$ correct of all the test
cases. The reason is that the last submission may be very close to the deadline
and for a student who is cheating, his/her submission can jump over the
submission which passes $75\%$ and goes to $100\%$ directly. By using the
submission passing $75\%$ test cases, we are able to create a more considerable variance in
the feature.

The experiment result for classifying students according to midterm grades is
shown in Table
\ref{table:submission_time_interval_dt_for_stu_categories_cm_midterm}. The
experiment result for classifying students according to their final exam grades
is shown in Table
\ref{table:submission_time_interval_dt_for_stu_categories_cm_final}.

\begin{table}[htb]
  \centering
  \begin{tabular}{cccc}
    \hline
    \hline
    PP&SP&GP&$\leftarrow$ classified as\\
    4               & 6                      &0              & PP\\
    5               &16                      &11             & SP\\
    5               &14                      &25             & GP\\
    \hline
  \end{tabular}
  \caption{Confusion Matrix for Submission Time Interval (midterm)}
  \label{table:submission_time_interval_dt_for_stu_categories_cm_midterm}
\end{table}
\begin{table}[htb]
  \centering
  \begin{tabular}{cccc}
    \hline
    \hline
    PP&SP&GP&$\leftarrow$ classified as\\
    5               & 6                      &2              & PP\\
    6               & 8                      &13             & SP\\
    2               &15                      &29             & GP\\
    \hline
  \end{tabular}
  \caption{Confusion Matrix for Submission Time Interval (final)}
  \label{table:submission_time_interval_dt_for_stu_categories_cm_final}
\end{table}

From table \ref{table:submission_time_interval_dt_for_stu_categories_cm_midterm}
and table \ref{table:submission_time_interval_dt_for_stu_categories_cm_final},
both models make some prediction for the poor-performance students. However, we
found that putting the submission time intervals into a linear-regression model can
make better predictions, and we will discuss it in the next section.

\subsection{Regression for Predicting the Grades}

This section describes the regression experiments for predicting the grades
aiming to explore the answer of the second research question: Can we predict
the students' numerical midterm grades and the final exam grades from the
students' behaviour?

\subsubsection{Time Interval between certain Submission and Deadline as the
  Feature}

The linear-regression model is based on the time interval between the time of the
submission that the student gets 75\% percent testcases passed of a given task
and the deadline.

\paragraph{Training Regression Model for Predicting Transformed Midterm Grades}

The tasks we used for regression to predict midterm grades only includes the
first 16 tasks (total 28 tasks) that were assigned before the midterm.

We used \texttt{stats.lm()} in R to build the linear-regression model and to
stabilize the variance we applied
\texttt{spreadLevelPlot()}~\cite{fox1997applied}\cite{hoaglin2003john} function
(It is a function for suggesting the numeric number for power transformation of
a linear-regression model) to find suggested power transformation.

The p-value for the model is $<2.2e-16$, which is much less than $0.05$ ($H_0$
hypothesis: no relationship between the time intervals and the transformed
midterm grades), which indicates there is a relationship between the time
intervals and the transformed midterm grades.

\begin{figure}[htb]
  \centering
  \includegraphics[width=\columnwidth]{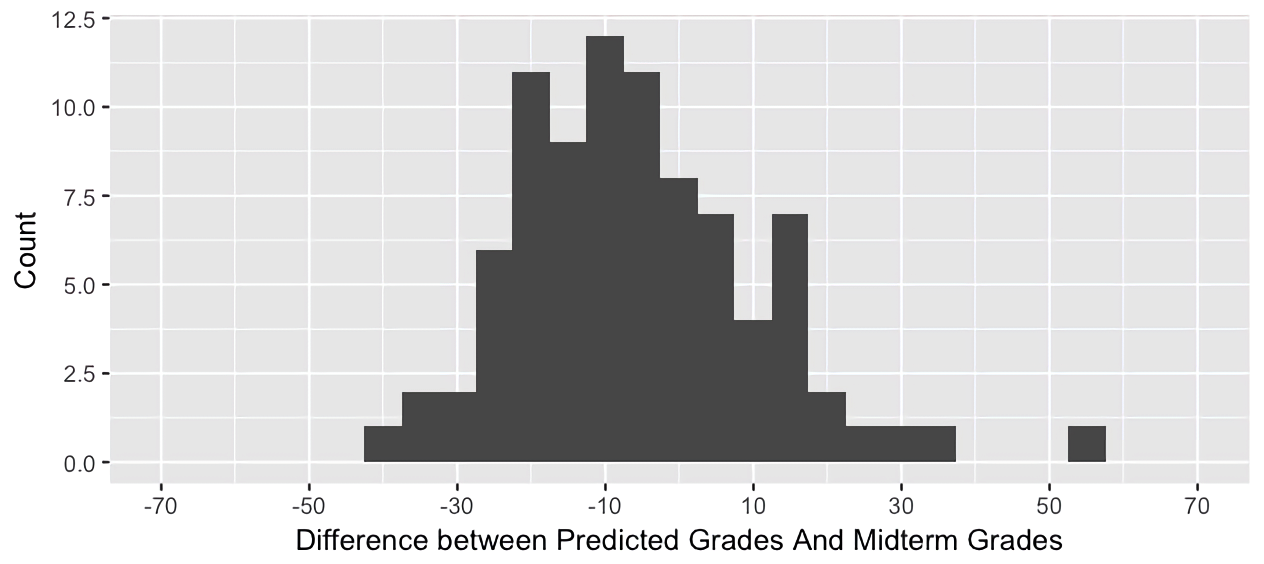}
  \caption{Histogram of Regression Difference between the Predicted grades and
    the Actual Grades}
  \label{fig:regression:histogram_midterm}
\end{figure}

Figure \ref{fig:regression:histogram_midterm} shows us the histogram of the
difference between the predicted grades and actual grades of the testing set. The
predicted midterm grades were calculated by applying reversal power
transformation. The mean of the difference between the predicted grades and the
actual midterm grades (maximum is $110$ points) is $-5.76$ points, and the
standard deviation is $16.44$ points.

\paragraph{Training Regression Model for Predicting Transformed Final Exam
  Grades}

The tasks we used for this experiment include all 28 tasks. The midterm grades
were not included. Similarly to transforming midterm grades, we also applied
\texttt{spreadLevelPlot()} function to find suggested power transformation in
order to stabilize the variance. The p-value for the model is $<2.2e-16$, which
indicates there is a relationship between the time intervals and the transformed
final exam grades.

\begin{figure}[htb]
  \centering
  \includegraphics[width=\columnwidth]{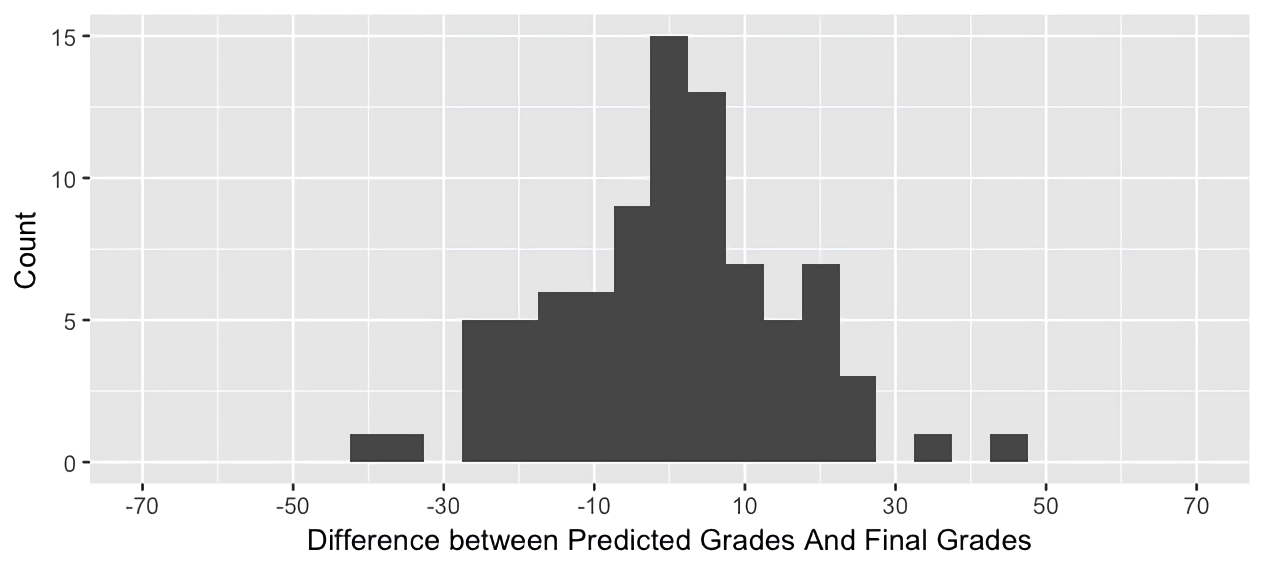}
  \caption{Histogram of Regression Difference between the Predicted and the
    Actual Grades}
  \label{fig:regression:histogram_final}
\end{figure}

Figure \ref{fig:regression:histogram_final} shows us the histogram of the
difference between the predicted grades and actual grades of the testing set. The
predicted final exam grades were calculated by applying reversal transformation.
The mean of the difference between the predicted grades and the actual final exam
grades (maximum is $120$ points) is $0.92$ points, and the standard deviation is
$17.12$ points.

In order to compare the performance of the regression model with the
classification models, we used the predicted the midterm grade or the final exam
grade from the regression models to generate the predicted categories and
compared them with the actual categories of students. The confusion matrix for
the midterm is shown in table \ref{fig:regression:regression_midterm_categories}, and
the confusion matrix for the final exam is shown in table
\ref{fig:regression:regression_final_categories}.

\begin{table}[htb]
  \centering
  \begin{tabular}{cccc}
    \hline
    \hline
    PP&SP&GP&$\leftarrow$ predicted as\\
    5               & 5                      &0              & PP\\
    5               &20                      &7              & SP\\
    0               &19                      &25             & GP\\
    \hline
  \end{tabular}
  \caption{Confusion Matrix from Predicted Midterm Grades}
  \label{fig:regression:regression_midterm_categories}
\end{table}
\begin{table}[htb]
  \centering
  \begin{tabular}{cccc}
    \hline
    \hline
    PP&SP&GP&$\leftarrow$ predicted as\\
    6               & 4                      &3              & PP\\
    1               &17                      &9              & SP\\
    1               &13                      &32             & GP\\
    \hline
  \end{tabular}
  \caption{Confusion Matrix from Predicted Final Exam Grades}
  \label{fig:regression:regression_final_categories}
\end{table}

From table \ref{fig:regression:regression_midterm_categories} and table
\ref{fig:regression:regression_final_categories}, we can observe that the
most significant number of each row and each column resides on the diagonal line from the
top-left corner to the bottom-right corner, meaning the model separates the
students well. Impressively, from the left-bottom zero and
top-right zero from table \ref{fig:regression:regression_midterm_categories}, it
indicates the midterm model successfully distinguishes poor-performance students
and good-performance students. Students from either category are not classified
into the other category.

\subsection{Comparison in terms of Poor-Performance Students}

We always care more about the poor-performance students comparing to others
because they are the students who need help urgently. Especially for
$1^{st}$ year engineering students, if they failed ECE 150, they are likely to
have much difficulty in passing other courses. Therefore, we want to compare the
results based on how models behave for poor-performance students (labelled as
positive).

Table \ref{fig:comparison:results_for_poor_performance_students_midterm} shows
us the results for poor-performance students on the midterm and table
\ref{fig:comparison:results_for_poor_performance_students_final} shows us the
results for poor-performance students on the final exam. Among them, what we
care about is the precision for different models. In this case, if any
students are classified as poor-performance students, they are likely to be in
trouble. However, by focusing on increasing the precision, we might put the
other poor-performance students who are not classified as poor-performance
students aside (low recall). The reason we let this situation happens is that
even if we cannot classify them, the worst case for them is that they remain in
a unchanged learning environment where they have to ask for help by themselves.
It is acceptable.

The importance of the values is $Precision > Fmeasure > Recall > FP\_rate$ ,
F-measure is a good way to judge the balance between precision and recall, so we
put it to the second place. FP\_rate comes the last because a model can easily
achieve a perfect FP\_rate by not making any positive instances. However, we
want our models to be able to make positive instances. Among them, precision,
F-measure and recall are the higher, the better while FP\_rate is the lower, the
better.

We refer to passing rate feature as \textbf{PR}, testcase outcomes as \textbf{TO},
Number of submissions as \textbf{NOS}, and submission time interval as
\textbf{STI} in the following tables.

\begin{table}[htb]
  \centering
  \begin{tabular}{cccccc}
    \hline
    \hline
    &Precision	&Recall   &F-Measure & FP\_rate	\\
    PR                                     	    &0.00	                    &0.00	                  &0.00              & \firstplace{0.03}\\
    TO\footnotemark &-	        &-   	    &-   & -   \\
    NOS                   	                    &0.20	                    &0.30	                  &0.24              & 0.16\\
    STI (Decision Tree)                         &\secondplace{0.29}       &\secondplace{0.40}     &\secondplace{0.33}& 0.13\\
    STI (Linear Regression) 	                  &\firstplace{0.50}	      &\firstplace{0.50}	    &\firstplace{0.50} & \secondplace{0.07}\\
    \hline
  \end{tabular}
  \caption{Results for Poor-Performance Students on the Midterm}
  \label{fig:comparison:results_for_poor_performance_students_midterm}
\end{table}
\footnotetext{The decision-tree model using testcase outcome as the feature
  predicts no student as a poor-performance student}

From table~\ref{fig:comparison:results_for_poor_performance_students_midterm},
we can see for the midterm, linear regression with submission time interval gives us
the best result for precision, F-measure and recall.

\begin{table}[htb]
  \centering
  \begin{tabular}{ccccc}
    \hline
    \hline
    &Precision	&Recall   &F-Measure & FP\_rate	\\
    PR                                     	    &0.40                 &0.15               &0.22              & \secondplace{0.04}\\
    TO                               	          &0.25                 &0.07               &0.12              & \secondplace{0.04}\\
    NOS                   	                    &\secondplace{0.50}   &\firstplace{0.54}  &\secondplace{0.52}& 0.10\\
    STI (Decision Tree)                         &0.39                 &0.39               &0.39              & 0.11\\
    STI (Linear Regression) 	                  &\firstplace{0.75}    &\secondplace{0.46} &\firstplace{0.57} & \firstplace{0.03}\\
    \hline
  \end{tabular}
  \caption{Results for Poor-Performance Students on the Final Exam}
  \label{fig:comparison:results_for_poor_performance_students_final}
\end{table}

From table~\ref{fig:comparison:results_for_poor_performance_students_final}, we
can see for the final exam, linear regression with submission time interval
gives us the best result for precision, F-measure, and FP\_rate. Only the recall
comes the second place.

\begin{figure}[!htb]
  \centering
  \includegraphics[width=\columnwidth]{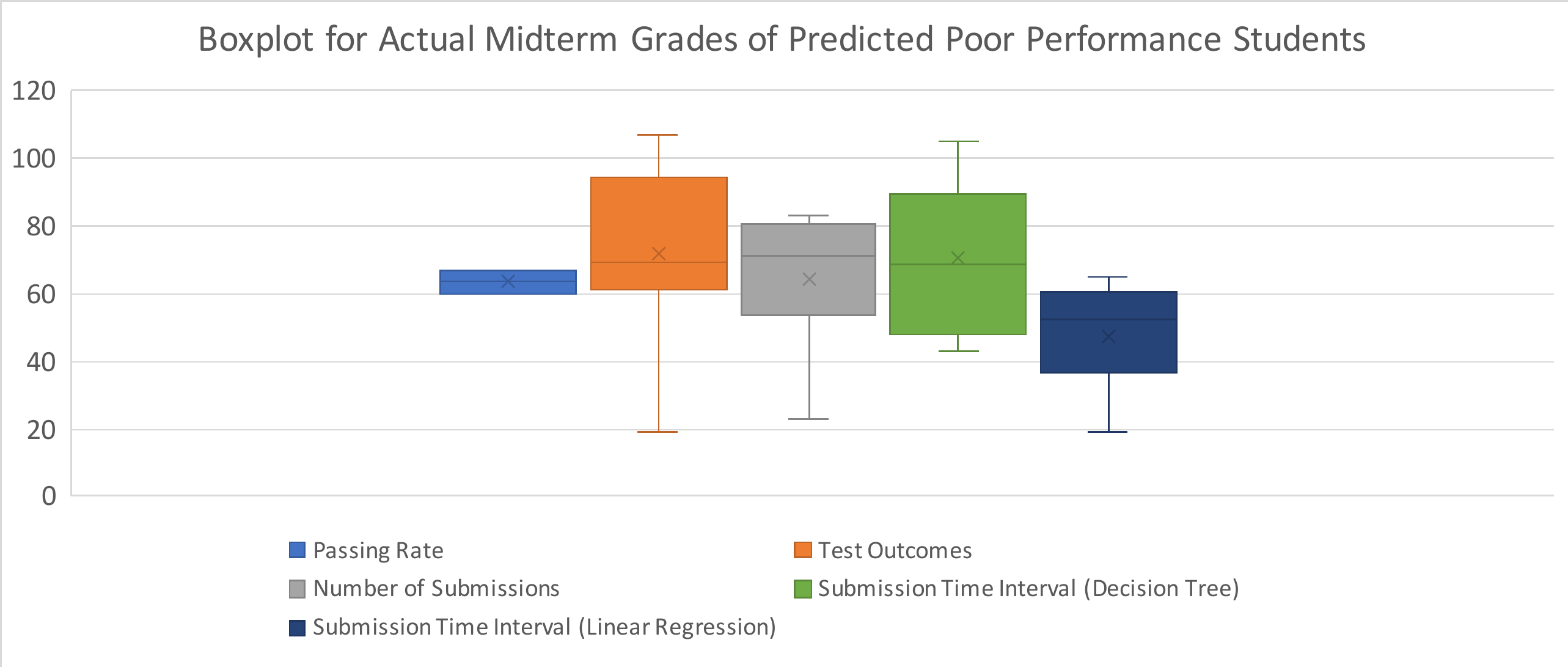}
  \caption{Boxplot for the Actual Midterm Grades of Predicted Poor Performance
    Students}
  \label{fig:comparison:boxplot_midterm}
\end{figure}

\begin{figure}[!htb]
  \centering
  \includegraphics[width=\columnwidth]{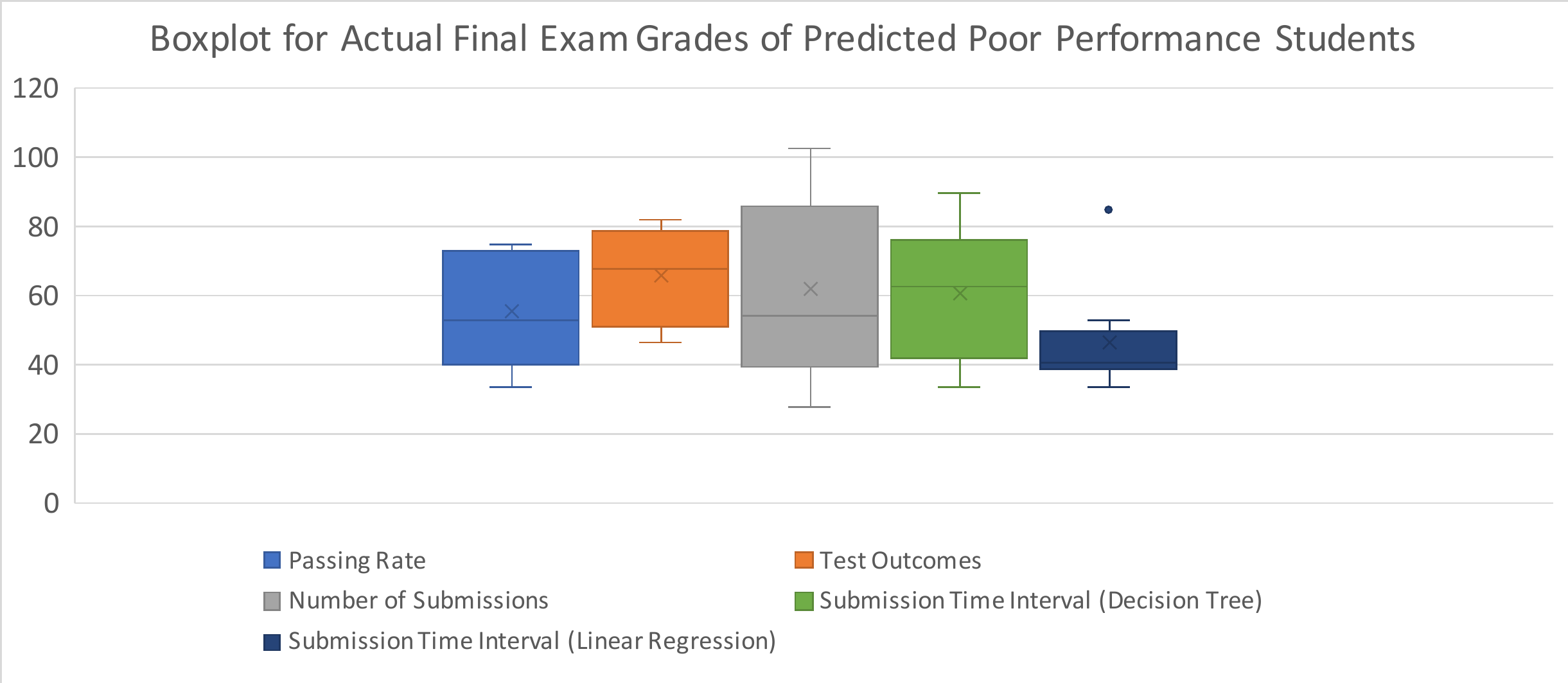}
  \caption{Boxplot for the Actual Final Exam Grades of Predicted Poor
    Performance Students}
  \label{fig:comparison:boxplot_final_without_midterm}
\end{figure}

In addition to those values, we also interested in what the actual grades are
for the predicted poor-performance students because although some students are
misclassified, their actual grades might be very close to poor-performance
students. Figure \ref{fig:comparison:boxplot_midterm} and figure
\ref{fig:comparison:boxplot_final_without_midterm} are showing the boxplots for
the actual grades of predicted poor-performance students.

From figure \ref{fig:comparison:boxplot_midterm}, we can see for the midterm, the
actual grades of predicted poor-performance students from submission time
linear-regression model are mostly the lowest among all. Also, we have a similar result
for the final exam, as shown in figure
\ref{fig:comparison:boxplot_final_without_midterm}.

Therefore, it is reasonable to say that the linear-regression model using the
submission time interval performs better than other models. However, for the
final exam, it seems the highest point for submission time interval (linear
regression) boxplot is an outlier. After we looked into the submission time
interval information of that specific student, we found out he/she submit
his/her code with an average of 23 hours before the deadline while the average
hours of other good-performance students reside mostly between 100 hours and 140
hours. Therefore, that student is not behaving like a good-performance student
in terms of the submission time interval feature, but rather a poor performance
student of whom the average hours reside mostly between 10 hours and 60 hours.
It makes sense that the student is predicted as poor performance student.

\subsection{Correlation between The Time Interval Information of an Assignment
  and Exam Grades}

This section is aiming to answer the third research question: Can we find any
interesting relations between the features generated (reflecting students'
behaviour), grades and student categories?. Currently, we have one finding that
the time interval of different assignment contributes to the prediction on the midterm
grades and the final exam grades differently.

We apply the linear-regression algorithm to different assignment (each
assignment contains several tasks with the same deadline) and calculate the
correlation coefficient for comparing the effect of the time interval
information from each assignment in order to have an insight into the correlation
between that and the exam grades. However, in all these experiments, no power
transformation is applied because the suggested power might not be constant across
the experiments. Thus, some underlying assumptions of the residuals of linear
regression are not met. The model is as:
$$grades=\beta_0+\sum\beta_ix_i$$
In the equation, the $i$ is between $1$ and the number of tasks of an assignment.
For a task, $x_i$ is the time interval between the submission time of a certain
submission passing $T\%$ testcases and the deadline. In our experiments, the
$T\%$ is set to 75\%. The difference from previous experiments is that 
only part of the time interval information of all tasks before
the midterm or the final exam is used. Specifically, we combined the time interval
information of tasks of an assignment to do the regression. For example, in
figure~\ref{fig:marmoset_graded_tasks_list}, assignment 1 contains 5 tasks. Then
in our experiment, we will use the time interval information of those 5 tasks
for prediction.

The dataset in the experiments is not split into a training set and a testing
set, but cross validation is applied. The values are averaged (The exact values
are lost from each split of the dataset during the cross validation. However,
the average value would give us a sense of the relation between each assignment
and midterm or the final exam grades).

\begin{table}[htb]
  \centering
  \begin{tabular}{|p{11em}|r|r|r|r|r|}
    \toprule
    \# assignment           & 0     & 1     & 2     & 3                  \\ 
    \midrule
    number of sub tasks     & 2     & 5     & 3     & 6                  \\ 
    \midrule
    Correlation coefficient & 0.37  & 0.46  & 0.51  & \firstplace{0.61}  \\
    \midrule
    Mean absolute error     & 17.09 & 16.02 & 15.53 & \firstplace{14.03} \\
    \midrule
    Root mean squared error & 20.94 & 19.99 & 19.35 & \firstplace{17.94} \\
    \bottomrule
  \end{tabular}
  \caption{Correlation between assignments (STI) and midterm grades}
  \label{table:contribution_for_midterm}%
\end{table}

\begin{table*}[!t]
  \begin{tabular}{|p{11em}|r|r|r|r|r|c|r|r|r|r|r|r|c|}
    \toprule
    \# assignment or midterm& 0     & 1     & 2     & 3       & midterm             & 4     & 5     & 6     & 7    \\
    \midrule
    number of sub tasks     & 2     & 5     & 3     & 6       & -                   & 3     & 3     & 4     & 2    \\
    \midrule
    Correlation coefficient & 0.36  & 0.39  & 0.47  & 0.56    & \firstplace{0.72}   & 0.54  & 0.55  & 0.52  & 0.57 \\
    \midrule
    Mean absolute error     & 15.22 & 15.09 & 14.61 & 13.36   & \firstplace{11.25}  & 13.57 & 13.58 & 13.78 & 13.23\\
    \midrule
    Root mean squared error & 18.93 & 18.73 & 17.98 & 16.81   & \firstplace{14.20}  & 17.10 & 16.97 & 17.35 & 16.67\\
    \bottomrule
  \end{tabular}
  \caption{Correlation between assignments (STI) or midterm grades and final
    exam grades}
  \label{table:contribution_for_final}
\end{table*}

Table \ref{table:contribution_for_midterm} shows that for the midterm, the
assignment immediately before the midterm has the most significant contribution to the midterm
grades (the most considerable correlation). It makes sense since there can be a large
portion of questions in the midterm asking about the content of that assignment,
if a student fails to manage time well, he/she may fail to fully understand that
assignment (they may still get a high score of that assignment since they can
ask for help from other students or cheat), which, in return, makes him/her
perform poorly in midterm.

\begin{figure}[htb]
  \centering
  \includegraphics[width=\columnwidth]{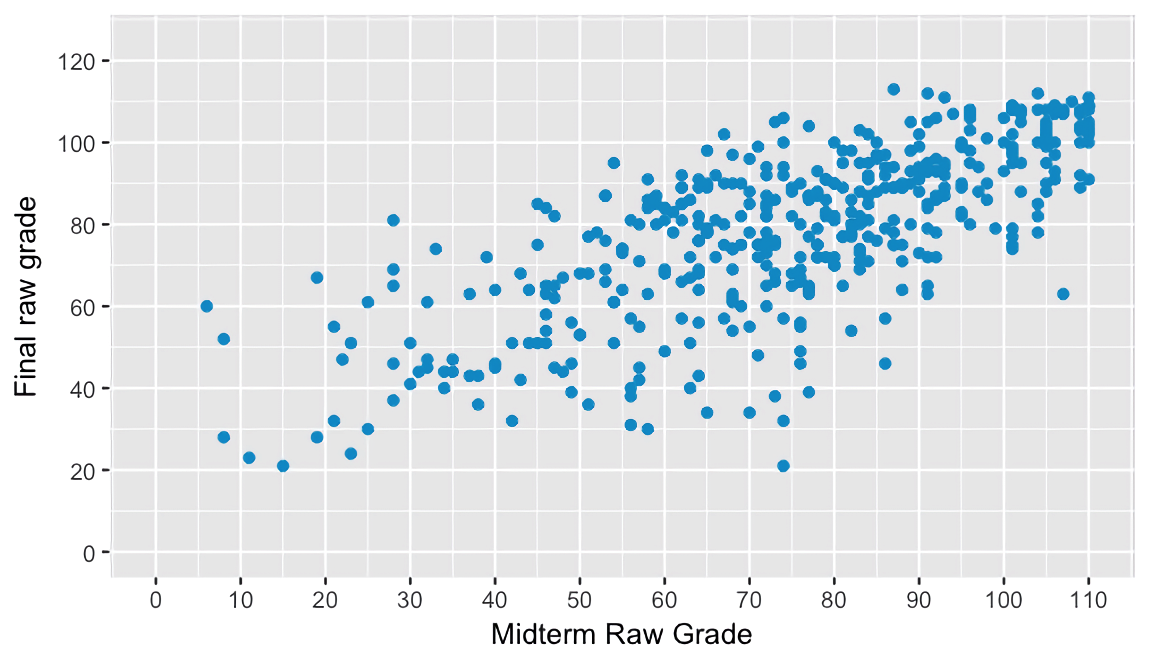}
  \caption{Relation between the midterm grades and the final exam grades}
  \label{fig:relation_midterm_and_final}
\end{figure}

Table \ref{table:contribution_for_final} show that for the final exam, it turns
out that the midterm grades has the most significant correlation with the final exam
grades. Figure \ref{fig:relation_midterm_and_final} shows us the scatter plot of
the midterm grades and the final exam grades. A linear relation can be observed.

\section{Conclusion and Future Work}
\label{sec:conclusion_and_future_work}

This paper is a summary of our preliminary study in predicting student
performance by using the data from the auto-grading system Marmoset. We carried
out two types of experiments: classification and regression, aiming to explore
the answer of three research questions.

\begin{enumerate}
\item If we put students into categories according to their final exam and
  midterm performances, can we create a model over the Marmoset data to
  understand the students' behaviour and predict those categories?
\item Can we predict the students' numerical midterm grades and the final exam
  grades using the Marmoset data?
\item Can we find any interesting relations between the Marmoset data, grades
  and student categories?
\end{enumerate}

For classification, we put students into categories according to their midterm
or final exam grades. The categories are good-performance students,
satisfactory-performance students and poor-performance students. We apply C4.5
decision-tree algorithm to the passing rate, testcases outcomes, the number of
submissions and submission time intervals, separately, and the results show
that, for predicting midterm performance, the classification models of passing
rate and testcases outcomes cannot label any poor-performance students
correctly. A strong bias exists tending to put every student into
satisfactory-performance students or good-performance students. For the final
exam, no such bias exists, and all the features are able to make limited
predictions.

For regression, we use linear regression algorithm on the time interval between
a student's first reasonable submission and the deadline (STI) to predict the
exam grades of the student. The results show that for the midterm, the mean of
the difference between predicted grades and actual midterm grades (maximum is
110 points) is -5.76 points, and the standard deviation is 16.44 points. For the
final exam, the mean of the difference between predicted grades and actual final
exam grades (maximum is 120 points) is 0.92 points, and the standard deviation
is 17.12 points.

In order to compare the results in terms of poor-performance students, we have
to convert the predicted grades from the linear-regression models into
categories. We categorize students by setting threshold on their predicted
grades. Then we calculate the FP\_rate, Precision, Recall and F-measure of the
poor-performance students for both cases. The importance of the values is
$Precision > Fmeasure > Recall > FP\_rate$. For both the midterm and the final
exam, we find out the regression model using the STI gives us the best value for
precision and F-measure. For students who are misclassified as poor-performance
students, we compare the boxplots of their actual grades from each model. We
find for both the midterm and the final exam, the linear-regression models using
the STI give the best result.

To compare the correlation between an individual assignment and exam grades, we
compared the contributions of the STI of different assignments, and it shows that for
the midterm, the assignment assigned right before the midterm exam has the
most significant contribution. For the final exam grades, the midterm grades contribute the most.

In summary, it is reasonable to say that the linear-regression model using
submission time interval performs better than other models in terms of
predicting poor-performance students and further researching on this might be
the best next step.

\bibliographystyle{ACM-Reference-Format} \bibliography{main}

\end{document}